\pdfoutput=1
\documentclass[sigconf]{acmart}

\usepackage{
    adjustbox,
    algorithm,
    amsfonts,
    amsmath,
    amsthm,
    booktabs,
    colortbl,
    diagbox,
    etoolbox,
    footnote,
    graphics,
    hyperref,
    makecell,
    microtype,
    multirow,
    nicefrac,
    pgfplots,
    pifont,
    schemabloc,
    setspace,
    siunitx,
    soul,
    subcaption,
    threeparttable,
    tikz,
    url,
    wrapfig,
    xcolor,
}

\usepackage[noend]{algpseudocode}
\usepackage[shortcuts]{extdash}
\usepackage[inline]{enumitem}

\usetikzlibrary{arrows,shapes,circuits.logic.US,shapes.gates.logic.IEC,calc,decorations.markings,patterns,matrix}
\tikzstyle{branch}=[fill,shape=circle,minimum size=3pt,inner sep=0pt]

\usepgfplotslibrary{groupplots}

\newcommand\func[2]{#1{\left(#2\right)}}
\newcommand\abs[1]{\left|{#1}\right|}

\newcommand{\dbtilde}[1]{\tilde{\raisebox{0pt}[0.85\height]{$\tilde{#1}$}}}
\usepackage{stackengine}
\stackMath
\newcommand\tsup[2][2]{%
 \def\useanchorwidth{T}%
  \ifnum#1>1%
    \stackon[-.5pt]{\tsup[\numexpr#1-1\relax]{#2}}{\scriptscriptstyle\sim}%
  \else%
    \stackon[.5pt]{#2}{\scriptscriptstyle\sim}%
  \fi%
}
	
\newcommand\circledone{\ding{172}}
\newcommand\circledtwo{\ding{173}}

\newcommand\mypatternone{north east lines}
\newcommand\mypatterntwo{north west lines}
\newcommand\mypatternthree{vertical lines}
\newcommand\mypatternfour{horizontal lines}
\newcommand\mycolourone{black!25!cyan}
\newcommand\mycolourtwo{black!25!magenta}
\newcommand\mycolourthree{red}
\newcommand\mycolourfour{black!25!green}
\newcommand\mycolourfive{blue}
\newcommand\mycoloursix{orange}

\usepackage{todonotes}
\newcommand\jd[1]{\todo[inline,backgroundcolor=pink]{JD: #1}}
\newcommand\ew[1]{\todo[inline,backgroundcolor=green!25]{EW: #1}}
\newcommand\ma[1]{\todo[inline,backgroundcolor=orange!25]{MA: #1}}
\newcommand\gc[1]{
    \textcolor{red}{[George: #1]}
}

    \renewcommand\jd[1]{}
    \renewcommand\ew[1]{}
    \renewcommand\gc[1]{}
    \renewcommand\ma[1]{}

\soulregister\ref{7}
\soulregister\subref{7}
\soulregister\cite{7}

\setcopyright{acmcopyright}
\copyrightyear{2022}
\acmYear{2022}
\acmDOI{10.1145/3490422.3502360}

\acmConference[FPGA '22]{Proceedings of the 2022 ACM/SIGDA International Symposium on Field-Programmable Gate Arrays}{February 27--March 1, 2022}{Virtual event.}
\acmBooktitle{Proceedings of the 2022 ACM/SIGDA International Symposium on Field-Programmable Gate Arrays (FPGA '22), February 27--March 1, 2022, Virtual event}
\acmPrice{15.00}
\acmISBN{978-1-4503-9149-8/22/02}

\begin{document}
\fancyhead{}

\title[Logic Shrinkage: Learned FPGA Netlist Sparsity for Efficient Neural Network Inference]{Logic Shrinkage: Learned FPGA Netlist Sparsity\\for Efficient Neural Network Inference}



\author{Erwei Wang}
\affiliation{%
    \institution{Imperial College London}
    \city{London}
    \country{United Kingdom}
}
\email{erwei.wang13@imperial.ac.uk}

\author{James J. Davis}
\affiliation{%
    \institution{Imperial College London}
    \city{London}
    \country{United Kingdom}
}
\email{james.davis@imperial.ac.uk}

\author{Georgios-Ilias Stavrou}
\affiliation{%
    \institution{Imperial College London}
    \city{London}
    \country{United Kingdom}
}
\email{georgios-ilias.stavrou18@imperial.ac.uk}

\author{Peter Y. K. Cheung}
\affiliation{%
    \institution{Imperial College London}
    \city{London}
    \country{United Kingdom}
}
\email{p.cheung@imperial.ac.uk}

\author{George A. Constantinides}
\affiliation{%
    \institution{Imperial College London}
    \city{London}
    \country{United Kingdom}
}
\email{g.constantinides@imperial.ac.uk}

\author{Mohamed S. Abdelfattah}
\affiliation{%
    \institution{Cornell University}
    \city{New York}
    \state{NY}
    \country{United States}
}
\email{mohamed@cornell.edu}

\renewcommand{\shortauthors}{Wang, et al.}

\begin{abstract}
    FPGA-specific DNN architectures using the native LUTs as independently trainable inference operators have been shown to achieve favorable area-accuracy and energy-accuracy tradeoffs.
    The first work in this area, LUTNet, exhibited state-of-the-art performance for standard DNN benchmarks.
    In this paper, we propose the learned optimization of such LUT-based topologies, resulting in higher-efficiency designs than via the direct use of off-the-shelf, hand-designed networks.
    Existing implementations of this class of architecture require the manual specification of the number of inputs per LUT, $K$.
    Choosing appropriate $K$ {\em a priori} is challenging, and doing so at even high granularity, {\em e.g.} per layer, is a time-consuming and error-prone process that leaves FPGAs' spatial flexibility underexploited.
    Furthermore, prior works see LUT inputs connected randomly, which does not guarantee a good choice of network topology.
    To address these issues, we propose {\em logic shrinkage}, a fine-grained netlist pruning methodology enabling $K$ to be automatically learned for every LUT in a neural network targeted for FPGA inference.
    By removing LUT inputs determined to be of low importance, our method increases the efficiency of the resultant accelerators.
    Our GPU-friendly solution to LUT input removal is capable of processing large topologies during their training with negligible slowdown.
    With logic shrinkage, we better the area and energy efficiency of the best-performing LUTNet implementation of the CNV network classifying CIFAR-10 by 1.54$\times$ and 1.31$\times$, respectively, while matching its accuracy.
    This implementation also reaches 2.71$\times$ the area efficiency of an equally accurate, heavily pruned BNN.
    On ImageNet with the Bi-Real Net architecture, employment of logic shrinkage results in a post-synthesis area reduction of 2.67$\times$ {\em vs} LUTNet, allowing for implementation that was previously impossible on today's largest FPGAs.
\end{abstract}

\ccsdesc[500]{Hardware~Reconfigurable logic and FPGAs}
\ccsdesc[500]{Hardware~Logic synthesis}
\ccsdesc[500]{Computing methodologies~Machine learning}

\keywords{LUT-based neural networks, binary neural networks, pruning.}


\maketitle

\section{Introduction}
\label{sec:intro}

    \begin{figure*}
	    \centering
	    \input{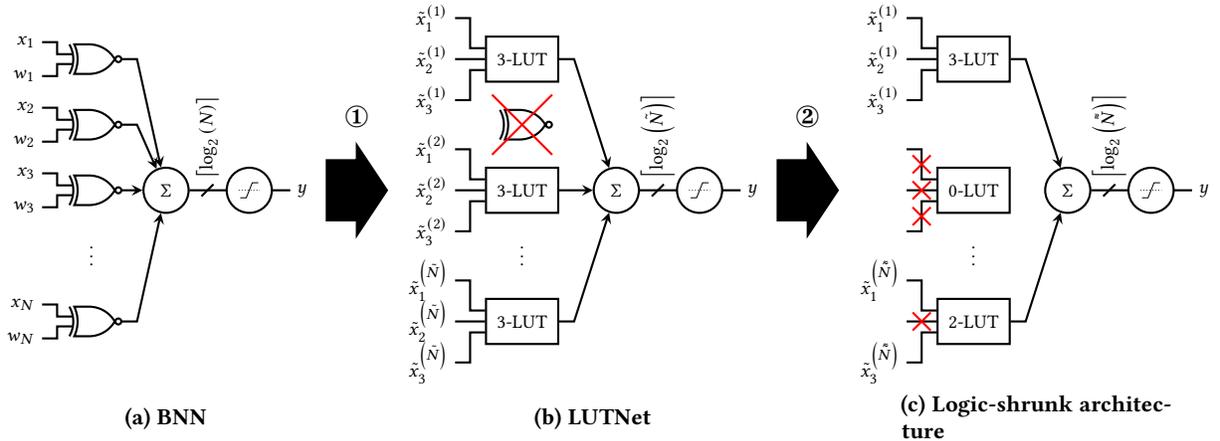}
	    \caption{
	        Summary of our proposed training regime, demonstrating the structural transformation of a single DNN channel from a BNN~(\subref{fig:ls_transform_bnn}) to a logic-shrunk architecture~(\subref{fig:ls_transform_lutnetls}).
	        In \circledone, the BNN is sparsified and logic-expanded, producing a LUTNet architecture~(\subref{fig:ls_transform_lutnet}) with $\tilde{N} \ll N$ $K$-LUTs (3-LUTs in this example) replacing $N$ XNORs.
	        LUT input pruning sees each $K$-LUT $n$ replaced with a $K^\prime_n$-LUT, where $K^\prime_n \leq K$.
	        When $K^\prime_n = 0$, LUT $n$ can be removed entirely.
	        This results in $\tsup{N} \leq \tilde{N}$.
        }
	    \label{fig:ls_transform}
	\end{figure*}

    Deep neural network (DNN) inference is particularly well suited to custom hardware acceleration due to the application's inherent parallelism.
    In order to exploit this in the quest for ever-greater performance within given area and power budgets, researchers and industrial practitioners alike are increasingly turning to low-precision data types~\cite{CSUR,SURV_EFFICIENT_DNN,SURV_CNN_ALGO}.
    Binary neural networks (BNNs), in which weights and activations assume one of just two values, see this concept taken to the extreme.
    Figure~\ref{fig:ls_transform_bnn} shows a generic BNN implementation of the quantized linear dot product operation central to DNN inference, wherein XNOR gates perform multiplication.
    Here, output $y = \func{\phi}{\boldsymbol{x}^\text{T}\boldsymbol{w}}$, with inputs $\boldsymbol{x} \in \left\{-1,1\right\}^N$, weights $\boldsymbol{w} \in \left\{-1,1\right\}^N$ and activation function $\phi : \mathbb{N}_{\geq 0} \to \left\{-1,1\right\}$.
    Such structures are compact and eminently parallelizable.
    When deployed on field-programmable gate arrays (FPGAs), however, their simplicity tends to lead to underuse of the rich compute and routing resources that the target device provides.
    
    We previously posited that more complex networks---netlists of small lookup tables (LUTs)---would ideally suit FPGA implementation due to their architectural similarity to the target fabric~\cite{LUTNET}.
    In that work, LUTNet, a BNN is first sparsified before its remaining XNORs are replaced with trainable $K:1$ Boolean operators: a process we termed {\em logic expansion}.
    Each of these, directly implementable as a $K$-LUT, has $K\times$ more inputs than its XNOR predecessor, enabling recovery of the accuracy lost due to pruning.
    Formally, LUT $n$ takes $\tilde{x}^{\left(n\right)}_i \sim \boldsymbol{x},~i \in \left\{1,\cdots,K\right\}$ as input.
    The weights are hardened within the LUT masks and so no longer appear externally.
    The result of this transformation is a fast and efficient task-specific inference accelerator.
    This is exemplified in Figure~\ref{fig:ls_transform_lutnet}, in which $\tilde{N}$ $K$-LUTs (here, 3-LUTs) have been substituted for $N$ XNOR gates.
    Since $\tilde{N} \ll N$, compaction of the adder tree more than compensates for the marginal area penalty attributable to the $K$-LUTs.
    With LUTNet, we reported area efficiency improvements of around 2$\times$ over ReBNet~\cite{BNN_CNN_REBNET_FCCM}, the state-of-the-art BNN at the time, for problems of widely varying scale.
    More recent tools, including NullaNet~\cite{nullanet} and LogicNets~\cite{logicnets}, also generate small LUTs as core components, but LUTNet remains unique in directly exposing a netlist's LUTs as differentiable functions trainable via stochastic gradient descent.
    
    In such a LUT-based network, fixed $K$ will inevitably be suboptimal.
    For example, while it may be the case that 6-LUTs map particularly well to a given device, $K=6$ may be too many (or too few) inputs for a given node to suit the training data.
    We therefore propose that the size of each LUT be {\em learned} during training.
    Starting from a netlist of $K$-LUTs, we achieve this by removing input connections determined to be unimportant, resulting in a new netlist in which $K^\prime_n \leq K~\forall n \in \left\{1,\cdots,\tilde{N}\right\}$.
    Where $K^\prime_n = 0$, LUT $n$ can be removed entirely.
    We exemplify this process in Figure~\ref{fig:ls_transform_lutnetls}, in which the total number of LUTs $\tsup{N} \leq \tilde{N}$, tending to further reduce area.
    The heterogeneity of the resultant netlists plays to the strengths of FPGA synthesis tools, which are adept at the low-level optimization of small Boolean functions.
    We find that networks constructed in this way are superior to their homogeneous counterparts, requiring fewer device resources to reach a target accuracy.
    
    We take inspiration from the field of neural architecture search (NAS), in which a sparse and efficient topology is typically found by cutting away parts of a dense network~\cite{SURV_NAS_CSUR}.
    While our end goal is similar, the netlist-level NAS we propose presents unique challenges.
    In particular, unlike standard topologies with a single weight per node, each node in a network of $K$-LUTs has $K$ inputs sharing $2^K$ trainable parameters.
    Severance of one LUT input requires the manipulation of all $2^K$ entries within the respective truth table.
    Given that modern DNNs contain hundreds of thousands or even millions of nodes, na\"ive operation on all of these would quickly become intractable.
    We thus present a vectorized implementation of our input pruning proposal ideally suited to GPU acceleration.
    
    In this paper, we present {\em logic shrinkage}: the automated search for, and construction of, DNN inference topologies featuring learned netlist sparsity.
    We make the following novel contributions.
    \begin{itemize}
        \item
            We propose a method for the evaluation of input connection salience within a netlist of LUTs used for DNN inference.
        \item
            We cast LUT input removal as a matrix-vector operation, enabling us to take advantage of GPUs for its realization.
        \item
            We present a TensorFlow-based implementation of logic shrinkage, in which DNNs composed of LUTs of fixed size are automatically transformed into sparser, heterogeneous networks more efficiently mappable onto FPGAs.
        \item
            We empirically explore the effects of logic shrinkage on area efficiency and accuracy via comparison with LUTNet~\cite{LUTNET}, our state-of-the-art FPGA-specific DNN inference topology, across a broad range of standard network models and datasets.
    	    We also experimentally determine logic shrinkage's impact on energy and training efficiency.
    	    Against LUTNet with fixed $K=4$, ordinarily the best-performing choice of constant $K$, we achieve area compression of 1.54$\times$ and an energy saving of 1.31$\times$ for the CNV network~\cite{BNN_CNN_FINN} classifying the CIFAR-10 dataset~\cite{CIFAR10} while reaching comparable accuracy.
    	    Finally, we report positive results at scale, with our logic-shrunk Bi-Real Net~\cite{birealnet} design classifying ImageNet~\cite{IMAGENET} demanding 2.67$\times$ lower post-synthesis area than LUTNet.
        \item
    	    We provide an open-source release\footnotemark of our work for the community to use and build upon.
    \end{itemize}
    \footnotetext{\url{https://github.com/awai54st/Logic-Shrinkage}}

\section{Related Work}

    \subsection{FPGA-Tailored DNN Architectures}
    \label{sec:related_fpga_spec_arch}
        
        LUT-based DNN inference accelerators have been shown to achieve remarkable performance when deployed on FPGAs.
        NullaNet~\cite{nullanet} and LogicNets~\cite{logicnets} were conceived with small-scale classification tasks in mind, for which they reached latency in the tens of nanoseconds and throughput in the hundreds of millions of samples per second.
        Going beyond FPGA-tailored network design, our previously proposed LUTNet topologies can be trained via stochastic gradient descent~\cite{LUTNET}.
        LUTNet's trainable netlists are compatible with common machine learning optimization strategies such as pruning, thereby affording opportunities for increased performance and efficiency.
        Furthermore, the LUTNet approach suits tasks spanning a broad range of scales, including ImageNet classification.
        
        LUTNet netlists tend to be large due to the one-to-one mapping between DNN nodes and LUTs.
        Consequently, in typical deployments, only a subset of network layers are logic-expanded: the remainder are kept as standard BNN structures.
        We have also proposed a time-multiplexed version of the LUTNet architecture, which negates the need for each LUT to be specific to a single node by reintroducing runtime-variable weights~\cite{LUTNET_TC}.
        This increases LUTNet's scalability, but also reduces its potential area and energy efficiency gains over BNNs due to the lost freedom in LUT specialization.
        
        We use LUTNet netlists as a starting point for logic shrinkage, and demonstrate that the resultant designs are more area and energy efficient.
        Our automated design flow maintains the deployment flexibility, scalability and ease of use of LUTNet's.
        To evaluate the potential of logic shrinkage in the most generic setting, we assume the use of hardened weights, in line with vanilla LUTNet.
        Our approach could be applied to time-multiplexed architectures, however we would similarly expect lower gains from doing so.
        
    \subsection{Activation Pruning}
    
        Activations within a DNN commonly contribute to its output to varying degrees.
        Activation pruning exploits this by assigning compute only to those with high relative importance (or salience), leading to increased efficiency.
        While crude attempts to establish salience, such as taking the mean of activations across a training dataset, were reportedly unsuccessful~\cite{PRU_CNN_SALIENCY_CRITERIA}, use of the partial derivative of a cost function with respect to the activations has been shown to work well~\cite{PRU_CNN_SALIENCY_CRITERIA,PRU_CNN_SNIP}.
        Such a partial derivative---an activation gradient---quantifies the impact of a perturbation of that activation on the output of the network.
        It is intuitive to therefore prioritize activations with low gradient magnitude for pruning.
        Molchanov~\emph{et al.}~\cite{PRU_CNN_SALIENCY_CRITERIA} and Lee~\emph{et al.}~\cite{PRU_CNN_SNIP} both took this approach, reporting state-of-the-art results with and without retraining, respectively.
        
        For the aforementioned works, which targeted standard DNN topologies, it was assumed that the gradients of activations within a layer are independent.
        This assumption does not hold in the context of LUT input pruning; input interdependence exists due to the configuration bits of each $K$-LUT since these are shared between each of its $K$ inputs.
        We introduce a pruning strategy that solves this problem, formulate it such that it is ideally suited to GPU acceleration and use it to generate area-efficient netlists.

    \subsection{Neural Architecture Search}
    
        Neural architecture search (NAS) automates the process of DNN design.
        In many NAS works, candidate functions are placed in parallel to form ``supernets'', after the training of which only those found to be of highest salience are retained.
        The granularity of functions that compose supernets varies.
        In DARTS, selections are made between small convolutional layers, with around 10 of these available to choose from in each instance~\cite{DARTS}.
        Candidate function outputs are scaled by trainable scaling factors before they are accumulated, making the search space continuous and therefore differentiable.
        The scaling factors capture function salience, and these are used post-training to determine the makeup of the final network.
        Works including DARTS have been shown to produce high-performance architectures orders of magnitude more quickly than their non-differentiable counterparts, including those using reinforcement learning~\cite{RL-based-NAS} and evolution~\cite{evolution-based-NAS}.
        The authors of AtomNAS proposed finer-grained search, decomposing convolutions into combinations of ``atomic blocks'' and greatly increasing the number of possible output architectures {\em vs} DARTS~\cite{ATOMNAS}.
        This richness in flexibility resulted in the production of state-of-the-art ImageNet classifiers.
        
        We propose a network topology search approach analogous to prior works on NAS.
        We start with an overprovisioned $K$-LUT-based architecture---a supernet---and selectively remove its redundancy at ultra-fine granularity via LUT input pruning.

\section{Background: Logic Expansion}\label{sec:background_lutnet}
	
	To enable post-logic expansion retraining for LUTNet, we defined an \emph{interpolating extension} to the complete set of $K:1$ Boolean operations as our training function~\cite{LUTNET}:
	\begin{equation}
        \func{f}{\tilde{\boldsymbol{x}}^{\left(n\right)}} = \sum_{\boldsymbol{d} \in \left\{-1,1\right\}^K}{\left(\hat{c}_{\boldsymbol{d}} \prod_{k = 1}^K{\left(1 - d_k\tilde{x}^{\left(n\right)}_k\right)}\right)}.
        \label{eqn:Lagrange_interpl}
    \end{equation}
	Real-valued parameters $\hat{\boldsymbol{c}}$ are trainable with stochastic gradient descent and, when binarized for use during inference, represent LUT masks, $\boldsymbol{c}$.
    \eqref{eqn:Lagrange_interpl} expands as
    \begin{equation}
        f{\left(\tilde{\boldsymbol{x}}^{\left(n\right)}\right)} =
        \begin{cases}
            \hat{c}_{\left(-1\right)}{\left(1 + \tilde{x}^{\left(n\right)}_1\right)} + \hat{c}_{\left(1\right)}{\left(1 - \tilde{x}^{\left(n\right)}_1\right)}	& \text{if}~K = 1\\
            \\
            \begin{aligned}[c]
                &\hat{c}_{\left(-1,-1\right)}{\left(1 + \tilde{x}^{\left(n\right)}_1\right)\left(1 + \tilde{x}^{\left(n\right)}_2\right)} \\
                &+ \hat{c}_{\left(-1,1\right)}{\left(1 + \tilde{x}^{\left(n\right)}_1\right)\left(1 - \tilde{x}^{\left(n\right)}_2\right)} \\
                &+ \hat{c}_{\left(1,-1\right)}{\left(1 - \tilde{x}^{\left(n\right)}_1\right)\left(1 + \tilde{x}^{\left(n\right)}_2\right)} \\
                &+ \hat{c}_{\left(1,1\right)}{\left(1 - \tilde{x}^{\left(n\right)}_1\right)\left(1 - \tilde{x}^{\left(n\right)}_2\right)}
            \end{aligned}	& \text{if}~K = 2\\
            \\
            \cdots  & \cdots
        \end{cases}
        \label{eqn:Lagrange_interpl_expand}
	\end{equation}
	for $K \in \mathbb{N}_{>0}$, with each polynomial comprising $2^K$ trainable parameters.
    We use a logic-expanded, retrained network as the starting point for logic shrinkage.
    
\section{Mechanics of Logic Shrinkage}\label{sec:logic_shrinkage}
    	
    \subsection{LUT Input Salience}
        
        The activation gradient-based salience criteria commonly used with standard neural networks are not directly applicable to netlist pruning due to the interdependence of LUT inputs.
        However, their fundamental concept---gauging an activation's importance by the impact on the network's outputs with respect to a change in that activation---remains relevant, thus we adopt it for the purpose of establishing LUT input salience.
        
        Consider a $K$-binary-input LUT with truth table entries encoded as $\left\{0,1\right\} \rightarrow \left\{-1,1\right\}$.
        Each entry represents the output with respect to a unique combination of inputs; changing one or more input values will alter the selection of LUT entry used as output.
        We define a particular LUT input's salience to be the sum of such changes across all combinations of the remaining inputs.
        If the flipping of a given input never leads to a change in LUT output, that input can clearly be removed without having any impact on the functionality of the network.
        Such an input therefore has zero salience.
        If toggling an input {\em sometimes}---but rarely---results in output change, we consider that input to be of low salience, while the opposite holds for an input whose toggling often causes the LUT's output to change.
        
        LUTNet-style Lagrangian interpolation, which we introduced to make LUTs differentiable~\cite{LUTNET}, presents us with an opportunity to more precisely quantify LUT input salience.
        Since LUT entries in this scenario are real-valued, output changes are typically less coarse than when operating in the binary domain.
        
        To exemplify our approach, Table~\ref{tab:and_gate_example} contains possible real-valued LUT entries $\hat{\boldsymbol{c}}$ of a 2-LUT, where $x_1$ and $x_2$ are its inputs.
        The LUT's entries will be binarized prior to synthesis; once this is done, this LUT will function as an AND gate.
        
        
        In Table~\ref{tab:and_gate_example}, the salience of input $x_1$, $s_1$, is defined as the total disturbance to the LUT output across both $x_2=1$ and $x_2=-1$ when $x_1$ experiences a change in sign, \emph{i.e.} the sum of column $\abs{\Delta_{x_1}}$.
        Similarly, the salience of $x_2$, $s_2$, is defined as the sum of row $\abs{\Delta_{x_2}}$.
        In general, we define the salience of $K$-LUT input $i$ as
        \begin{equation}
            s_i = \sum_{\boldsymbol{d}_1 \in \left\{-1,1\right\}^{i-1}}\sum_{\boldsymbol{d}_2 \in \left\{-1,1\right\}^{K-i}}{\abs{\hat{c}_{\left(\boldsymbol{d}_1,1,\boldsymbol{d}_2\right)} - \hat{c}_{\left(\boldsymbol{d}_1,-1,\boldsymbol{d}_2\right)}}}.
            \label{eqn:salience_metric}
        \end{equation}
        
        From Table~\ref{tab:and_gate_example}, since $s_1 > s_2$, we can conclude that toggles of input $x_1$ lead to greater impact on the LUT output than toggles of $x_2$. $x_2$ is therefore less important than $x_1$ and so should be prioritized for disconnection.
        Once the less-salient inputs of a network's LUTs have been identified, we can turn to their removal.
        
        We experimented with other candidate salience criteria---including weight gradient-~\cite{PRU_CNN_SNIP} and Taylor expansion-based~\cite{PRU_CNN_SALIENCY_CRITERIA} methods---before settling on the aforedescribed approach.
        While these were shown to work well for conventional DNN node pruning, we did not observe positive results in their use for LUT input removal.

    \subsection{Pruning}
    	
    	In a similar vein to the establishment of salience, LUT input pruning also requires a nonstandard approach.
    	With a conventional neural network node, an activation can be removed by setting its corresponding weight to zero.
    	The removal of an input from a $K$-LUT, on the other hand, requires the manipulation of all of the $2^K$ $\hat{\boldsymbol{c}}$ parameters that define the contents of the LUT.
    	
    	Here we demonstrate the process of removing LUT inputs using the 2-LUT in \eqref{eqn:Lagrange_interpl_expand} as an example.
    	In order to remove input $\tilde{x}^{\left(n\right)}_1$, the LUT mask $\hat{\boldsymbol{c}}$ should be transformed into $\hat{\boldsymbol{c}}^{\prime}$ such that $\hat{c}_{\left(-1,-1\right)}^{\prime} = \hat{c}_{\left(1,-1\right)}^{\prime}$ and $\hat{c}_{\left(-1,1\right)}^{\prime} = \hat{c}_{\left(1,1\right)}^{\prime}$.
    	Countless functions can be used to achieve this.
    	Of them, we chose the computationally cheapest: assignment using the means of their pre-shrinkage values.
    	The removal of input $\tilde{x}^{\left(n\right)}_1$ is thus achieved by performing
    	\begin{equation}
            \begin{aligned}[c]
    	    \hat{c}_{\left(-1,-1\right)}^{\prime} &= \nicefrac{1}{2}\left(\hat{c}_{\left(-1,-1\right)} + \hat{c}_{\left(1,-1\right)}\right)\\
    	    \hat{c}_{\left(-1,1\right)}^{\prime} &= \nicefrac{1}{2}\left(\hat{c}_{\left(-1,1\right)} + \hat{c}_{\left(1,1\right)}\right)\\
    	    \hat{c}_{\left(1,-1\right)}^{\prime} &= \nicefrac{1}{2}\left(\hat{c}_{\left(-1,-1\right)} + \hat{c}_{\left(1,-1\right)}\right)\\
    	    \hat{c}_{\left(1,1\right)}^{\prime} &= \nicefrac{1}{2}\left(\hat{c}_{\left(-1,1\right)} + \hat{c}_{\left(1,1\right)}\right)
            \end{aligned}
        \label{eqn:input_removal_2lut_example_remove_x1}
    	\end{equation}
    	Similarly, the removal of LUT input $\tilde{x}^{\left(n\right)}_2$ is achieved as
    	\begin{equation}
            \begin{aligned}[c]
    	    \hat{c}_{\left(-1,-1\right)}^{\prime} &= \nicefrac{1}{2}\left(\hat{c}_{\left(-1,-1\right)} + \hat{c}_{\left(-1,1\right)}\right)\\
    	    \hat{c}_{\left(-1,1\right)}^{\prime} &= \nicefrac{1}{2}\left(\hat{c}_{\left(-1,-1\right)} + \hat{c}_{\left(-1,1\right)}\right)\\
    	    \hat{c}_{\left(1,-1\right)}^{\prime} &= \nicefrac{1}{2}\left(\hat{c}_{\left(1,-1\right)} + \hat{c}_{\left(1,1\right)}\right)\\
    	    \hat{c}_{\left(1,1\right)}^{\prime} &= \nicefrac{1}{2}\left(\hat{c}_{\left(1,-1\right)} + \hat{c}_{\left(1,1\right)}\right) 
            \end{aligned}
            \label{eqn:input_removal_2lut_example_remove_x2}
    	\end{equation}
    	
    	Referring back to the 2-LUT example in Table~\ref{tab:and_gate_example}, removal of less-salient input $x_2$ requires the application of \eqref{eqn:input_removal_2lut_example_remove_x2} to the LUT mask, $\hat{\boldsymbol{c}}$.
    	This results in new parameters $\hat{c}_{\left(-1,-1\right)}^{\prime} = \hat{c}_{\left(-1,1\right)}^{\prime} = -0.88$ and $\hat{c}_{\left(1,-1\right)}^{\prime} = \hat{c}_{\left(1,1\right)}^{\prime} = 0.02$.
    	Once $\hat{\boldsymbol{c}}^\prime$ is binarized, the 2-LUT performs the single-input function $y = x_1$, \emph{i.e.} it is transformed into a wire.
	
    \begin{table}
        \centering
        \caption{An example 2-LUT truth table with real-valued entries, in this case representing an AND gate. $\Delta_{x_i}$ captures the change in LUT output with respect to a change in input $x_i$.}
        \begin{tabular}{c|cc|c} 
            \diagbox{$x_2$}{$x_1$} & -1 & 1 & $\abs{\Delta_{x_1}}$\\ 
            \hline
            -1 & -0.90 & -0.01 & \bf{0.89}\\ 
            1 & -0.85 & 0.05 & \bf{0.90}\\ 
            \hline
            $\abs{\Delta_{x_2}}$ & \bf{0.05} & \bf{0.06} &\\
        \end{tabular}
        \label{tab:and_gate_example}
    \end{table}
    
    \subsection{Pruning at Scale}
    	
    	While pruning when $K=2$, as exemplified in~\eqref{eqn:input_removal_2lut_example_remove_x1} and~\eqref{eqn:input_removal_2lut_example_remove_x2}, is straightforward, the complexity of these operations increases exponentially with $K$.
    	Logic shrinkage of one $K$-LUT involves the transformation of $2^K$ parameters, and the assignments are unique for each of the $\sum^K_{i=1}{K \choose i} = 2^K-1$ possible LUT input combinations.
    	This complexity further scales with the number of LUTs being trained.
    	To ensure scalability, the implementation of our pruning method must therefore take advantage of the high-performance linear algebraic capabilities of modern GPUs and DNN training frameworks.
    	
    	We implement functions such as \eqref{eqn:input_removal_2lut_example_remove_x1} and \eqref{eqn:input_removal_2lut_example_remove_x2} as matrix-vector multiplications $\hat{\boldsymbol{c}}^{\prime} = \boldsymbol{U}\hat{\boldsymbol{c}}$ with a transformation matrix $\boldsymbol{U} \in \mathbb{R}^{2^K\times2^K}$.
    	Continuing with those examples, the removal of input $\tilde{x}^{\left(n\right)}_1$ in \eqref{eqn:input_removal_2lut_example_remove_x1} and of input $\tilde{x}^{\left(n\right)}_2$ in \eqref{eqn:input_removal_2lut_example_remove_x2} are performed as
	    \begin{gather*}
    	    \begin{pmatrix}
    	            \hat{c}_{\left(-1,-1\right)}^{\prime}\\ 
    	            \hat{c}_{\left(-1,1\right)}^{\prime}\\ 
    	            \hat{c}_{\left(1,-1\right)}^{\prime}\\ 
    	            \hat{c}_{\left(1,1\right)}^{\prime}
    	    \end{pmatrix} = \frac{1}{2}
    	    \begin{pmatrix}
    	            1 & 1 & 0 & 0\\
    	            1 & 1 & 0 & 0\\
    	            0 & 0 & 1 & 1\\
    	            0 & 0 & 1 & 1 
    	    \end{pmatrix}
    	    \begin{pmatrix}
    	            \hat{c}_{\left(-1,-1\right)}\\ 
    	            \hat{c}_{\left(-1,1\right)}\\ 
    	            \hat{c}_{\left(1,-1\right)}\\ 
    	            \hat{c}_{\left(1,1\right)}
    	    \end{pmatrix},\\
    	    \text{\em i.e.}\quad\boldsymbol{U}_1 = \frac{1}{2} \boldsymbol{I}^{2\times2} \otimes \boldsymbol{1}^{2\times2}
    	\end{gather*}
    	and
    	\begin{gather*}
    	    \begin{pmatrix}
    	            \hat{c}_{\left(-1,-1\right)}^{\prime}\\ 
    	            \hat{c}_{\left(-1,1\right)}^{\prime}\\ 
    	            \hat{c}_{\left(1,-1\right)}^{\prime}\\ 
    	            \hat{c}_{\left(1,1\right)}^{\prime}
    	    \end{pmatrix} = \frac{1}{2}
    	    \begin{pmatrix}
    	            1 & 0 & 1 & 0\\
    	            0 & 1 & 0 & 1\\
    	            1 & 0 & 1 & 0\\
    	            0 & 1 & 0 & 1 
    	    \end{pmatrix}
    	    \begin{pmatrix}
    	            \hat{c}_{\left(-1,-1\right)}\\ 
    	            \hat{c}_{\left(-1,1\right)}\\ 
    	            \hat{c}_{\left(1,-1\right)}\\ 
    	            \hat{c}_{\left(1,1\right)}
    	    \end{pmatrix},\\
    	    \text{\em i.e.}\quad\boldsymbol{U}_2 = \frac{1}{2} \boldsymbol{1}^{2\times2} \otimes \boldsymbol{I}^{2\times2},
    	\end{gather*}
    	respectively, where $\otimes$ is the Kronecker product and use of $\boldsymbol{U}_i$ causes the removal of LUT input $i$.
    	
    	The removal of a single input can be conceptualized as the merging of LUT parameter pairs followed by the forking of their means back to their original locations.
    	This is achieved by $\boldsymbol{1}^{2\times2}$ in the aforementioned examples.
    	The Kronecker product with the identity matrix permutes the merges and forks as required.
    	In general,
    	\begin{equation}
    	    \boldsymbol{U}_i = \frac{1}{2} \boldsymbol{I}^{2^{K-i}\times2^{K-i}} \otimes \boldsymbol{1}^{2\times2} \otimes \boldsymbol{I}^{2^{i-1}\times2^{i-1}} .
    	    \label{eqn:U_i}
    	\end{equation}
    	Where removal of multiple LUT inputs is desired, $\boldsymbol{U}_i$ for each input $i$ can simply be multiplied together to form a single transformation matrix, $\boldsymbol{U}$, before application.
    	
    	The construction of $\boldsymbol{U}$, although computationally expensive, is a one-time process that we have found to never exceed 10~s.
    	During retraining, logic shrinkage is implemented as one instance of matrix-vector multiplication, which is ideally suited to GPU acceleration.
    	
    	Although a post-shrinkage LUT mask $\hat{\boldsymbol{c}}^\prime$ will always represent a simpler function, dependent on fewer inputs, than its predecessor $\hat{\boldsymbol{c}}$, $\hat{\boldsymbol{c}}^\prime$ will retain $2^K$ parameters.
    	While this means that a post-shrinkage netlist will contain redundancy, a benefit of this is that such a netlist will remain compatible with the existing LUTNet implementation flow.
    	Our experiments revealed that Vivado effectively recognizes and removes this redundancy during synthesis with no noticeable overhead.
    	Representation of sparse input connections in a dense format, as we propose, also simplifies our training software.
    	
    \subsection{Iterative Pruning}
    
        \begin{algorithm}[t]
    	    \caption{Logic shrinkage retraining process.}
    	    \begin{flushleft}
        	    \hspace*{\algorithmicindent} \textbf{Inputs:}\\
        	    \hspace*{\algorithmicindent}\hspace*{\algorithmicindent} $K \in \mathbb{N}$, \Comment{\# pre-shrinkage inputs per LUT}\\
        	    \hspace*{\algorithmicindent}\hspace*{\algorithmicindent} $\tilde{N} \in \mathbb{N}$, \Comment{\# pre-shrinkage LUTs}\\
        	    \hspace*{\algorithmicindent}\hspace*{\algorithmicindent} $\delta \in \left[0,1\right]$, \Comment{Target sparsity}\\
        	    \hspace*{\algorithmicindent}\hspace*{\algorithmicindent} $T \in \mathbb{N}$, \Comment{\# shrinkage iterations}\\
        	    \hspace*{\algorithmicindent}\hspace*{\algorithmicindent} $P \in \mathbb{N}$, \Comment{\# epochs per shrinkage iteration}\\
        	    \hspace*{\algorithmicindent}\hspace*{\algorithmicindent} $\hat{\boldsymbol{C}} \in \mathbb{R}^{2^K\times \tilde{N}}$ \Comment{Pre-shrinkage LUT masks}\\
                \hspace*{\algorithmicindent} \textbf{Output:}\\
        	    \hspace*{\algorithmicindent}\hspace*{\algorithmicindent} $\hat{\boldsymbol{C}}^{\prime} \in \mathbb{R}^{2^K\times \tilde{N}}$ \Comment{Post-shrinkage LUT masks}
    	    \end{flushleft}
    	    \begin{algorithmic}[1]
    	    \Procedure{logicShrink}{}
    	        \State $\hat{\boldsymbol{C}}^{\prime} \leftarrow \hat{\boldsymbol{C}}$
                \For{$t \leftarrow \left\{1,\cdots,T\right\}$}
                    \State $\boldsymbol{S} \leftarrow \func{\text{getsalience}}{\hat{\boldsymbol{C}}^{\prime}}$ \Comment{Per \eqref{eqn:salience_metric}; $\boldsymbol{S} \in \mathbb{R}^{\tilde{N} \times K}_{\geq 0}$}
                    \State $\boldsymbol{r} \leftarrow \func{\text{getRankOrder}}{\func{\text{vec}}{\boldsymbol{S}}}$
                    \State $\delta_t \leftarrow \delta \times \nicefrac{t}{T}$
                    \State $\boldsymbol{M} \leftarrow \func{\text{vec}^{-1}_{\tilde{N} \times K}}{\boldsymbol{1}_{\boldsymbol{r} < \delta_t\tilde{N}K}}$ \Comment{$\boldsymbol{M} \in \left\{0,1\right\}^{\tilde{N} \times K}$}
                    \State $\boldsymbol{V} \leftarrow \boldsymbol{0}^{\tilde{N}\times 2^K \times 2^K}$
                    \For{$n \leftarrow \left\{1,\cdots,\tilde{N}\right\}$}
                        \State $\boldsymbol{V}_n \leftarrow \boldsymbol{I}^{2^K \times 2^K}$
                        \For{$i \leftarrow \left\{1,\cdots,K\right\}$}
                            \If{$m_{ni} = 1$}
                                \State $\boldsymbol{V}_n \leftarrow \boldsymbol{V}_n \boldsymbol{U}_i$ \Comment{Per \eqref{eqn:U_i}; $\boldsymbol{V}_n \in \mathbb{R}^{2^K\times 2^K}$}
                            \EndIf
                        \EndFor
                    \EndFor
                    \State $\hat{\boldsymbol{C}}^\prime \leftarrow \func{\text{retrain}}{\hat{\boldsymbol{C}}^\prime, \boldsymbol{V}, \text{epochs}=P}$
                \EndFor
                \State \Return $\hat{\boldsymbol{C}}^\prime$
            \EndProcedure
    	    \end{algorithmic}   
    	    \label{alg:logic_shrinkage}
    	\end{algorithm}
        
        The authors of many network pruning works, including Han~\emph{et al.}~\cite{PRU_CNN_TRAIN_PRUNE_RETRAIN} and See~\emph{et al.}~\cite{PRU_LSTM_NMT_TRAIN_PRUNE_RETRAIN}, proposed pruning across multiple iterations, with each including a post-pruning retraining phase.
        In keeping with this approach, we separate our LUT input pruning process into multiple iterations, each greedier than the last, with retraining following each.
        In early experiments, we confirmed that this approach outperforms one-shot pruning, and found that $T=3$ iterations with $P=20$ retraining epochs following each performed favorably.
        As exemplified in Figure~\ref{plot:training_curves}, this setup results in training stability being reached quickly in each iteration.
        
        Algorithm~\ref{alg:logic_shrinkage} details the iterative logic shrinkage training process.
        In each of the $T$ total iterations, salience scores of all LUT inputs in the subset of the network subject to logic shrinkage are evaluated using \eqref{eqn:salience_metric} and then ranked.
        The input sparsity for iteration $t$, $\delta_t$, increases with $t$ until the target sparsity $\delta$ has been reached.
        Binary mask $\boldsymbol{M}$ indicates the low-salience LUT inputs to be pruned.
        Finally, logic shrinkage transformation matrices $\boldsymbol{U}$ are constructed based on $\boldsymbol{M}$, and the network is retrained with input connections sparsified for $P$ epochs.
        When retraining, we consistently apply all $\boldsymbol{U}$s formed in order to ensure that inputs previously severed by logic shrinkage remain so from then on.
        The topology of the portion of the network not subject to logic shrinkage is preserved throughout this process, but its parameters remain trainable.

\section{Evaluation}

    \subsection{Implementation}
    
        \begin{figure}
    	    \centering
    	    \begin{tikzpicture}[label distance=2mm,decoration={markings,mark= at position 0.5 with{\node[font=\footnotesize] {/};} },scale=0.8, every node/.style={scale=0.8}]

    \tikzstyle{ItemBox} = [rectangle, draw, text centered, minimum height=1em, minimum width=10em, node distance=3cm, text width=9em]
    \tikzstyle{ActionCircle} = [rounded rectangle, draw, text centered, minimum height=1em, minimum width=10em, node distance=3cm, text width=9em]
    \tikzstyle{TextBox} = [rectangle, text centered, minimum height=1em, minimum width=10em, node distance=3cm, text width=9em]
    
    \node[ItemBox] at (0,6) (model) {Model, dataset, activation precision};
    \node[ItemBox] at ($(model)+(0,-1.3)$) (pruning_threshold) {Node sparsity $\theta$};
    
    \node[ActionCircle] at ($(model)+(4.3,0)$) (training) {Training};
    \node[ActionCircle] at ($(pruning_threshold)+(4.3,0)$) (pruning) {Node pruning};
    \node[ActionCircle] at ($(pruning)+(0,-0.8)$) (logic_expansion) {Logic expansion};
    
    \node[ActionCircle, red] at ($(logic_expansion)+(0,-0.8)$) (logic_shrinkage) {Logic shrinkage};
    \node[ItemBox, red] at ($(logic_shrinkage)+(-4.3,0)$) (act_pruning_threshold) {LUT input sparsity $\delta$};
    
    \node[ItemBox] at ($(logic_shrinkage)+(0,-1)$) (trained_lutnet) {Trained network};
    
    \node[ActionCircle] at ($(trained_lutnet)+(0,-1)$) (lutnet_synflow) {LUTNet implementation flow};
    
    
    
    
    
    \draw [->] (model.east) -- (training.west);
    \draw [->] (pruning_threshold.east) -- (pruning.west);
    \draw [->, red] (act_pruning_threshold.east) -- (logic_shrinkage.west);
    \draw [->] (training.south) -- (pruning.north);
    \draw [->] (pruning.south) -- (logic_expansion.north);
    \draw [->, red] (logic_expansion.south) -- (logic_shrinkage.north);
    \draw [->, red] (logic_shrinkage.south) -- (trained_lutnet.north);
    
    \draw [->] (trained_lutnet.south) -- (lutnet_synflow.north);
    
    
    
    
    \draw [dashed] ($(training.north)+(-2.1,0.8)$) -- ($(training.north)+(2.1,0.8)$) -- ($(logic_shrinkage.south)+(2.1,-0.2)$) -- ($(logic_shrinkage.south)+(-2.1,-0.2)$) -- ($(training.north)+(-2.1,0.8)$);
    
    \draw [dashed] ($(lutnet_synflow.north)+(-2.3,0.2)$) -- ($(lutnet_synflow.north)+(2.3,0.2)$) -- ($(lutnet_synflow.south)+(2.3,-0.8)$) -- ($(lutnet_synflow.south)+(-2.3,-0.8)$) -- ($(lutnet_synflow.north)+(-2.3,0.2)$);
    
    
    \node[TextBox] at ($(training.north)+(0,0.4)$) {TensorFlow};
    \node[TextBox, text width=15em] at ($(lutnet_synflow.south)+(0,-0.4)$) {Vivado, LUTNet RTL gnerator};

\end{tikzpicture}
        	\caption{Incorporation of logic shrinkage within LUTNet's fully automated training and FPGA implementation flow.}
        	\label{fig:impl}
        \end{figure}
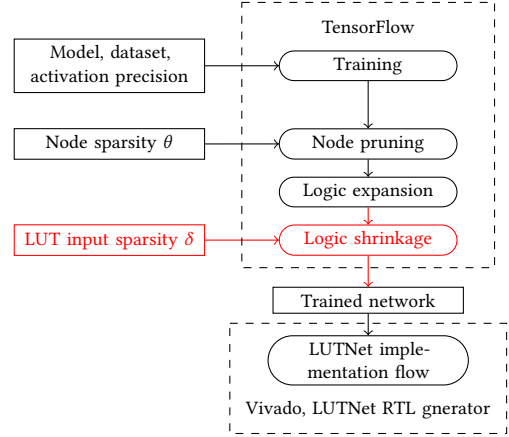

        For ease of development and evaluation, we engineered logic shrinkage as a bolt-on addition to the existing LUTNet training and hardware implementation flow~\cite{LUTNET}.
        A high-level view of the augmented flow, with the logic shrinkage stage annotated in red, can be found in Figure~\ref{fig:impl}.
        Now, in addition to the network model, training dataset, input precision and node pruning level that LUTNet takes as input, the user provides their desired LUT input pruning level as well.
        The back-end FPGA implementation steps remain unchanged.
        
        In common with LUTNet, employment of logic shrinkage necessitates no FPGA knowledge.
        Parameterized Keras layers and C++ templates are provided for training and implementation, respectively, enabling low-effort construction of dataflow DNN engines.

    \subsection{Benchmarks}

        \begin{table*}
    		\centering
    		\caption{
    		    Network architectures for evaluated benchmarks.
    		    Conv\textsubscript{$x,y,z$} denotes a convolutional layer with $x$ outputs, kernel size $y \times y$ and stride $z$.
    		    FConn\textsubscript{$x$} is a fully connected layer with $x$ outputs.
    		    MaxPool\textsubscript{$x,y$} is an $x \times x$ maximum-pooling layer with stride $y$, and BatchNorm and SoftMax are batch normalization and normalized exponential layers, respectively.
    		    ResBlk\textsubscript{$x,y,z$} denotes a residual block with two Conv\textsubscript{$x,y,z$} layers, each followed by a BatchNorm.
    		    Layers in bold were unrolled and targeted for logic expansion (and shrinkage).
    		    For ImageNet, the residual block in bold had its first convolutional layer unrolled and targeted.
            }
		    \begin{tabular}{ccc}
				\toprule
				Dataset & Model	& Network architecture\\
				\midrule
				\multirow{2}{*}{MNIST~\cite{MNIST}}  & \multirow{2}{*}{LFC~\cite{BNN_CNN_FINN}}  & FConn\textsubscript{256}, BatchNorm, \textbf{FConn\textsubscript{256}}, BatchNorm, \textbf{FConn\textsubscript{256}}, BatchNorm,  \textbf{FConn\textsubscript{256}}, BatchNorm,\\ & & \textbf{FConn\textsubscript{10}}, BatchNorm, SoftMax\\
				\midrule
				\multirow{3}{*}{\shortstack{SVHN~\cite{SVHN} \&\\CIFAR-10~\cite{CIFAR10}}} & \multirow{3}{*}{CNV~\cite{BNN_CNN_FINN}} & Conv\textsubscript{64, 3, 1}, BatchNorm, Conv\textsubscript{64, 3, 1}, BatchNorm, MaxPool\textsubscript{2, 2}, Conv\textsubscript{128, 3, 1},  BatchNorm,	Conv\textsubscript{128, 3, 1},\\ & & BatchNorm, MaxPool\textsubscript{2, 2}, Conv\textsubscript{256, 3, 1}, BatchNorm, \textbf{Conv\textsubscript{256, 3, 1}}, BatchNorm, FConn\textsubscript{512}, BatchNorm,\\ & & FConn\textsubscript{512}, BatchNorm, FConn\textsubscript{10}, BatchNorm, SoftMax\\
				\midrule
				\multirow{2}{*}{ImageNet~\cite{IMAGENET}} & \multirow{2}{*}{Bi-Real-18~\cite{birealnet}} & Conv\textsubscript{64, 7, 2}, BatchNorm, MaxPool\textsubscript{3, 2}, ResBlk\textsubscript{64, 3, 1}, ResBlk\textsubscript{64, 3, 1}, ResBlk\textsubscript{128, 3, 2}, ResBlk\textsubscript{128, 3, 2},\\ & & ResBlk\textsubscript{256, 3, 2}, \textbf{ResBlk\textsubscript{256, 3, 2}}, ResBlk\textsubscript{512, 3, 2}, ResBlk\textsubscript{512, 3, 2}, FConn\textsubscript{10}, SoftMax\\
				\bottomrule
			\end{tabular}
			\label{tab:model_info}
		\end{table*}
        
        We evaluated our approach using the DNN model and dataset combinations detailed in Table~\ref{tab:model_info}.
    	Hardware implementations for all datasets other than ImageNet targeted the Xilinx Kintex UltraScale XCKU115.
    	For ImageNet, we targeted the largest FPGA available to us: the Virtex UltraScale+ XCVU9P.
    	All implementations met timing at 200~MHz.
        Our primary comparison point was LUTNet, trained as we described in its original publication~\cite{LUTNET}.
        Where possible, we also maintained the BNN baseline, ReBNet~\cite{BNN_CNN_REBNET_FCCM}, used as the starting point for LUTNet's logic expansion, and considered its test accuracy to be a performance floor.
        
        For fairness of comparison to vanilla LUTNet (and ReBNet), we used identical experimental settings to those employed for its evaluation with MNIST, SVHN and CIFAR-10~\cite{LUTNET}.
        Implementations for these datasets included all layers: those selected for logic expansion (and subsequent shrinkage) were unrolled, with the remainder left identical to the BNN starting point.
        For ImageNet, our design encompassed the target layer only due to the complexity of implementing the remaining layers.
        In all cases, layers selected for logic expansion and shrinkage are marked in bold in Table~{\ref{tab:model_info}}.

    \subsection{Training Specifics}
    
        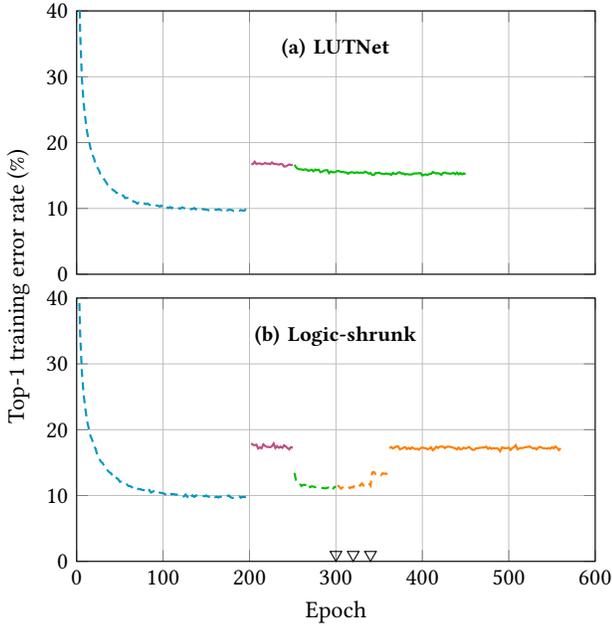
\begin{figure}
            \begin{tikzpicture}
    
    \begin{groupplot} [
		width=\columnwidth,
		height=0.6\columnwidth,
		group style={group size=1 by 2, xlabels at=edge bottom, xticklabels at=edge bottom, vertical sep=1em},
		ymin=0,
		ymax=40,
		xmax=600,
		xmin=0,
		xlabel near ticks,
		xlabel={Epoch},
        ylabel near ticks,
        grid=both,
        grid style={line width=.1pt, draw=gray!10},
        major grid style={line width=.2pt,draw=gray!50},
	]
        
        \nextgroupplot
            
        \addplot [thick, \mycolourone, densely dashed] table [y=training_error, x=epoch] {data/cifar_training_curve_ls_baseline/train.txt}; \label{plt:cifar_training_curve_baseline_train}
        \addplot [thick, \mycolourtwo] table [y=training_error, x=epoch] {data/cifar_training_curve_ls_baseline/prune.txt}; \label{plt:cifar_training_curve_baseline_prune}
        \addplot [thick, \mycolourfour] table [y=training_error, x=epoch] {data/cifar_training_curve_ls_baseline/retrain.txt}; \label{plt:cifar_training_curve_baseline_retrain}
        
        \node [text width=10em, anchor=north] at (axis description cs:0.5, 1) {\subcaption{LUTNet\label{plot:cifar_training_curve_baseline}}};
            
        \nextgroupplot [
            ylabel={Top-1 training error rate (\%)},
            every axis y label/.append style={at=(ticklabel cs:1), xshift=0.5em}
        ]
            
        \addplot [thick, \mycolourone, densely dashed] table [y=training_error, x=epoch] {data/cifar_training_curve/train.txt}; \label{plt:cifar_training_curve_train}
        \addplot [thick, \mycolourtwo] table [y=training_error, x=epoch] {data/cifar_training_curve/prune.txt}; \label{plt:cifar_training_curve_prune}
        \addplot [thick, \mycolourfour, densely dashed] table [y=training_error, x=epoch] {data/cifar_training_curve/le_retrain.txt}; \label{plt:cifar_training_curve_le_retrain}
        \addplot [thick, \mycoloursix, densely dashed] table [y=training_error, x=epoch] {data/cifar_training_curve/ls_retrain.txt}; \label{plt:cifar_training_curve_ls_retrain}
        \addplot [thick, \mycoloursix] table [y=training_error, x=epoch] {data/cifar_training_curve/ls_retrain_bin.txt}; \label{plt:cifar_training_curve_ls_retrain_bin}
        
        \addplot [mark=triangle,
            mark size=2.5pt,
            mark options={
                draw=black,
                fill=white,
            },
            only marks,
            every mark/.append style={rotate=180},
        ]
        table {%
        300 1
        320 1
        340 1
        }; \label{plt:ls_epochs}
        
        \node [text width=10em, anchor=north] at (axis description cs:0.5, 1) {\subcaption{Logic-shrunk\label{plot:cifar_training_curve}}};
    
    \end{groupplot}

\end{tikzpicture}
        	\caption{
        	    Training error for CNV classifying CIFAR-10 using LUTNet~(\subref{plot:cifar_training_curve_baseline}) and logic-shrunk~(\subref{plot:cifar_training_curve}) architectures, during initial training~(\ref{plt:cifar_training_curve_train}), post-node pruning retraining~(\ref{plt:cifar_training_curve_prune}), post-logic expansion retraining~(\ref{plt:cifar_training_curve_baseline_retrain}) and post-logic shrinkage retraining~(\ref{plt:cifar_training_curve_ls_retrain}).
        	    Phases with binarized forward propagation are denoted with solid lines; those with high-precision ({\tt float32}) forward propagation are shown dashed.
        	    Annotations~(\ref{plt:ls_epochs}) mark epochs at which logic shrinkage was applied.
            }
        	\label{plot:training_curves}
        \end{figure}
        
        \subsubsection{Small-Scale Datasets}
        \label{sec:training_specifics_small}
        
            For our experiments with MNIST~\cite{MNIST}, SVHN~\cite{SVHN} and CIFAR-10, pretrained ReBNet BNNs were first node-pruned and logic-expanded following the LUTNet approach (described in Section~\ref{sec:background_lutnet}) before being logic-shrunk (Section~\ref{sec:logic_shrinkage}).
            We inserted four new retraining phases between the post-node pruning~(\ref{plt:cifar_training_curve_prune}) and post-logic expansion~(\ref{plt:cifar_training_curve_baseline_retrain}) phases performed for LUTNet shown in Figure~\ref{plot:cifar_training_curve_baseline}.
            These are reflected in Figure~\ref{plot:cifar_training_curve}.
            After logic expansion, we performed 50 epochs of retraining with high-precision forward propagation~(\ref{plt:cifar_training_curve_le_retrain}), with a further 20~(\ref{plt:cifar_training_curve_ls_retrain}) performed following each of three logic shrinkage iterations.
            Finally, 200 epochs with binarized forward propagation~(\ref{plt:cifar_training_curve_ls_retrain_bin}) were performed, matching the final phase of LUTNet training.
            We chose these numbers of epochs and logic-shrinkage iterations since, as exemplified in Figure~\ref{plot:training_curves}, training accuracy saturation was achieved at or before the end of each phase.
            All training phases were executed in TensorFlow and accelerated using Nvidia RTX~3090 GPUs.
        
        \subsubsection{ImageNet}
        
            We also experimented with the ImageNet dataset.
            For this task, we prepared a pretrained Bi-Real Net model~\cite{birealnet}, Bi-Real-18, as our starting point, and then performed the retraining process outlined in Section~\ref{sec:training_specifics_small}.
            Here, we ran post-logic expansion retraining for 32 epochs (rather than 50), post-logic shrinkage retraining for eight epochs per iteration (rather than 20) and final, binarized retraining for 64 epochs (rather than 200).
            These numbers were again chosen due to our observance of accuracy stability.

    \subsection{Area Efficiency}
        
        In line with the prior FPGA-tailored DNN works detailed in Section~\ref{sec:related_fpga_spec_arch}, our primary objective was to maximize the area efficiency of our implementations.
        We define this as the number of device LUTs required to construct a network able to achieve a particular test accuracy for a given dataset while operating at a given classification rate.
        In all of our experiments, throughput remained fixed, thus we need only consider area {\em vs} accuracy.
    
        \subsubsection{Pruning Sparsity Tuning}
            
            \begin{figure*}
                \centering
                \input{plots/err_area_groupplot_points_only}
            	\caption{
            	    Area-accuracy tradeoff for LUTNet~(\ref{plt:tradeoff_2lutnet_pointsonly_baseline_1}~\ref{plt:tradeoff_2lutnet_pointsonly_baseline_2}~\ref{plt:tradeoff_2lutnet_pointsonly_baseline_3}~\ref{plt:tradeoff_2lutnet_pointsonly_baseline_4}~\ref{plt:tradeoff_2lutnet_pointsonly_baseline_5}) and logic-shrunk~(\ref{plt:tradeoff_2lutnet_pointsonly_curve_1}~\ref{plt:tradeoff_2lutnet_pointsonly_curve_2}~\ref{plt:tradeoff_2lutnet_pointsonly_curve_3}~\ref{plt:tradeoff_2lutnet_pointsonly_curve_4}~\ref{plt:tradeoff_2lutnet_pointsonly_curve_5}) implementations of the CNV network classifying the CIFAR-10 dataset with (initial) LUT size $K =$~ 2~(\subref{plot:tradeoff_2lutnet_pointsonly}), 4~(\subref{plot:tradeoff_4lutnet_pointsonly}) and 5~(\subref{plot:tradeoff_5lutnet_pointsonly}).
            	    Each color/shape reflects a distinct node sparsity $\theta$.
            	    Along a given curve, each logic-shrunk point is representative of a different LUT input sparsity $\delta$.
            	    The reference accuracy---that for unpruned ReBNet---is annotated on each $y$-axis~(\ref{plt:ref_acc}).
                }
            	\label{plot:AREA_LUT_TRADEOFF_GROUPPLOT_POINTS_ONLY}
            \end{figure*}
            
            We began by seeking to understand the interplay between the sparsity afforded to us through BNN node sparsification (by tuning $\theta$) and LUT input pruning ($\delta$).
            To this end, Figure~\ref{plot:AREA_LUT_TRADEOFF_GROUPPLOT_POINTS_ONLY} shows the achieved whole-network area {\em vs} top-1 test accuracy for LUTNet and logic-shrunk implementations of the CNV network trained to classify the CIFAR-10 dataset.
            Each point marks the mean of five differently seeded training runs, with an error bar indicating its range.
            For reference, the mean test error rate of ReBNet without pruning---again averaged over five training runs---is also shown~(\ref{plt:ref_acc}).
            Filled markers~(\ref{plt:tradeoff_2lutnet_pointsonly_baseline_1} \ref{plt:tradeoff_2lutnet_pointsonly_baseline_2} \ref{plt:tradeoff_2lutnet_pointsonly_baseline_3} \ref{plt:tradeoff_2lutnet_pointsonly_baseline_4} \ref{plt:tradeoff_2lutnet_pointsonly_baseline_5}) reflect results for LUTNet, split into those with LUT size $K=$~2~(Figure~\ref{plot:tradeoff_2lutnet_pointsonly}), 4~(\ref{plot:tradeoff_4lutnet_pointsonly}) and 5~(\ref{plot:tradeoff_5lutnet_pointsonly}).
            Each color/shape represents a distinct node sparsity $\theta$.
            Unfilled markers~(\ref{plt:tradeoff_2lutnet_pointsonly_curve_1} \ref{plt:tradeoff_2lutnet_pointsonly_curve_2} \ref{plt:tradeoff_2lutnet_pointsonly_curve_3} \ref{plt:tradeoff_2lutnet_pointsonly_curve_4} \ref{plt:tradeoff_2lutnet_pointsonly_curve_5}) capture area {\em vs} accuracy for logic-shrunk implementations with varying LUT input sparsity $\delta$.
            Along each colored line, implementations all had the same $K$ and $\theta$, varying only in $\delta$.
            Logic-shrunk designs used the respective fixed-$K$ LUTNet architecture as the starting point for logic shrinkage, after which they contained LUTs up to size $K$.
    		
    		By comparing across data points of different shapes/colors, one can clearly observe that the error rate increases as more aggressive node pruning is applied.
    		This trend is consistent across both the LUTNet and logic-shrunk implementations.
    		Figure~\ref{plot:AREA_LUT_TRADEOFF_GROUPPLOT_POINTS_ONLY} also reveals relatively consistent area-accuracy tradeoffs exposed through the variance of LUT input sparsity $\delta$ for each combination of $K$ and $\theta$.
    		As $\delta$ increases, connection pruning becomes more aggressive, pushing data points to the left.
            The error rate decreases at first due to the removal of redundant logic from the netlist.
            Beyond each curve's inflection point, the pruning becomes too harsh; we thus begin to see the error rate rise.
    		Also notice that, in some cases, $K$-LUT-based implementations outperform unpruned ReBNet (660196 LUTs) despite occupying as little as a quarter of its area.
    		This speaks to the increased expressiveness of these architectures over BNNs.
            
    		Inspection of Figures~\ref{plot:tradeoff_2lutnet_pointsonly} and \ref{plot:tradeoff_4lutnet_pointsonly} reveals that, in some cases, logic-shrunk implementations consume more area than the LUTNet architectures they were shrunk from.
    		This is counterintuitive since logic shrinkage reduces netlist complexity by severing LUT connections; it never adds them.
    		We attribute this effect, which is more pronounced in denser networks (higher $\theta$) of smaller LUTs (lower $K$), to Vivado's heuristic-based placement and routing algorithms.
    		
    		These experiments suggest that the performance of logic-shrunk networks is more sensitive to the tuning of node sparsity $\theta$ than LUT input sparsity $\delta$.
    		Figure~\ref{plot:AREA_LUT_TRADEOFF_GROUPPLOT_POINTS_ONLY} contains design points with $\theta$ ranging from 91.0 to 98.0\% and $\delta$ in the range 0.0--87.5\%.
    		We can see that a 7~pp change in $\theta$ has a larger impact on area-accuracy behavior than a change in $\delta$ more than 10$\times$ in magnitude.
    		We thus recommend that $\theta$ be fine-tuned with $\delta=0$ prior to increasing $\delta$ with fixed $\theta$.
    		We have found $\delta = 75\%$ to be a reasonable starting point.
    
        \subsubsection{Pareto-Optimality Analysis}
        
            \begin{figure*}
                \centering
                \input{plots/err_area_groupplot}
            	\caption{
            	    Pareto-optimal frontiers of the LUTNet~(\ref{plt:tradeoff_2lutnet_baseline_1}) and logic-shrunk~(\ref{plt:tradeoff_2lutnet_curve_1}) data points from Figure~\ref{plot:AREA_LUT_TRADEOFF_GROUPPLOT_POINTS_ONLY}, with LUTNet~(\ref{plt:2lutnet_pareto}) frontiers shown alongside those for logic-shrunk implementations with (initial) LUT size $K =$~2~(\subref{plot:tradeoff_2lutnet},~\ref{plt:ls_2lutnet_pareto}), 4~(\subref{plot:tradeoff_4lutnet},~\ref{plt:ls_4lutnet_pareto}) and 5~(\subref{plot:tradeoff_5lutnet},~\ref{plt:ls_5lutnet_pareto}).
            	    Pruned ReBNet data~(\ref{plt:tradeoff_bnn}~\ref{plt:bnn_pareto}) are also present.
            	    All frontiers are overlaid in (\subref{plot:tradeoff_pareto_fronts}).
            	    Arrows indicate the area decrease between the best-performing LUTNet and logic-shrunk implementations with accuracy bounded within $\pm$0.3~pp of unpruned ReBNet's~(\ref{plt:ref_acc}).
                }
            	\label{plot:AREA_LUT_TRADEOFF_GROUPPLOT}
            \end{figure*}
            
            \begin{figure}
            	\begin{tikzpicture}
    \begin{axis}[
        ybar,
        width=\columnwidth,
        height=0.65\columnwidth,
		enlarge x limits=0.3,
        ymin=0,
        ymax=420000,
		ytick scale label code/.code={\pgfmathparse{int(#1)}$\text{Area occupancy (LUTs)} \cdot 10^{\pgfmathresult}$},
        every y tick scale label/.style={at=(yticklabel cs:0.5), rotate=90, anchor=south},
        xtick=data,
        xticklabels from table={data/energy_efficiency.txt}{Name},
        xtick align=outside,
        /pgf/bar width=15pt,
        legend image code/.code={
            \draw[#1, bar width=6pt, yshift=-0.3em] plot coordinates {(0cm,0.8em)};
        }
    ]
        
        \addplot [bar shift=-10pt, thick, pattern=\mypatternone, pattern color=\mycolourone] table [x=id, y=post_place] {data/energy_efficiency.txt};
        \label{plt:post_place}

    \end{axis}
    
    \begin{axis}[
        ybar,
        width=\columnwidth,
        height=0.65\columnwidth,
		enlarge x limits=0.3,
        ymin=0,
        ymax=420000,
        axis x line=none,
        axis y line=none,
        xtick=\empty,
		ytick=\empty,
        /pgf/bar width=15pt,
        legend image code/.code={
            \draw[#1, bar width=6pt, yshift=-0.3em] plot coordinates {(0cm,0.8em)};
        }
    ]
        
        \addplot [bar shift=10pt, thick, pattern=\mypatterntwo, pattern color=\mycolourtwo] table [x=id, y=post_place] {data/energy_efficiency_ls.txt}
        node [pos=0, rotate=90, anchor=west, yshift=-10pt] {1.34$\times$}
        node [pos=0.5, rotate=90, anchor=west, yshift=-10pt] {1.54$\times$}
        node [pos=1, rotate=90, anchor=west, yshift=-10pt] {1.76$\times$};
        \label{plt:post_place_ls}
    
    \end{axis}

\end{tikzpicture}
            	\caption{
            	    Post-implementation LUT requirements of the best-performing LUTNet~(\ref{plt:post_place}) and logic-shrunk~(\ref{plt:post_place_ls}) implementations from Figure~\ref{plot:AREA_LUT_TRADEOFF_GROUPPLOT_POINTS_ONLY} with accuracy $\pm$0.3~pp from that of unpruned ReBNet.
            	    Annotations indicate LUT decreases.
                }
            	\label{plot:area_savings}
            \end{figure}
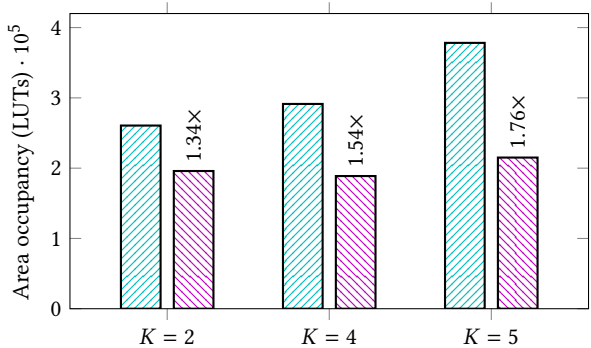
            
            Figures~\ref{plot:tradeoff_2lutnet}--\ref{plot:tradeoff_5lutnet} feature the data points taken from Figures~\ref{plot:tradeoff_2lutnet_pointsonly}--\ref{plot:tradeoff_5lutnet_pointsonly} with the addition of Pareto-optimal frontiers for the LUTNet~(\ref{plt:tradeoff_2lutnet_baseline_1}~\ref{plt:2lutnet_pareto}) and logic-shrunk~(\ref{plt:tradeoff_2lutnet_curve_1}~\hspace{-1.5mm}\begin{tabular}{c}\makecell{\ref{plt:ls_2lutnet_pareto}\vspace{-3.3mm}\\\ref{plt:ls_4lutnet_pareto}\vspace{-3.3mm}\\\ref{plt:ls_5lutnet_pareto}}\end{tabular}\hspace{-1.5mm}) implementations with identical (initial) LUT size $K$.
            For reference, points for the pruned ReBNet implementations used as starting points for logic expansion are also included~(\ref{plt:tradeoff_bnn}~\ref{plt:bnn_pareto}).
            From these plots, we can quickly establish that logic shrinkage facilitates a significant area improvement---savings of up to 1.76$\times$ while remaining bounded within $\pm$0.3~pp of the unpruned ReBNet accuracy---over LUTNet.
            As $K$ increases, the area gap between LUTNet and logic-shrunk designs increases, indicating that netlists of fixed-$K$-LUTs with higher $K$ are more redundant.
            Since logic shrinkage removes this redundancy, we would expect implementations with differing initial $K$ reaching comparable accuracy to be similar in size.
            We explore this hypothesis in Figure~\ref{plot:area_savings}, in which the pairs of Pareto-optimal LUTNet and logic-shrunk implementations that resulted in the savings marked by dashed lines in Figure~\ref{plot:AREA_LUT_TRADEOFF_GROUPPLOT} are featured.
            As expected, the area of the logic-shrunk designs is relatively stable.
            
            In Figure~\ref{plot:tradeoff_pareto_fronts}, we overlay the frontiers across all $K$ taken from Figures~\ref{plot:tradeoff_2lutnet}--\ref{plot:tradeoff_5lutnet}.
            The LUTNet frontier~(\ref{plt:2lutnet_pareto}) in Figure~\ref{plot:tradeoff_pareto_fronts} captures all Pareto-optimal LUTNet points from the preceding subfigures.
            In comparison to that for pruned ReBNet~(\ref{plt:bnn_pareto}), its placement demonstrates the significant area efficiency gain when moving from XNOR- to LUT-based networks for deployment on FPGAs.
            However, with logic shrinkage, we go further: all three logic-shrunk frontiers reflect improvement over LUTNet, with that using $K=4$ as the starting point~(\ref{plt:ls_4lutnet_pareto}) performing the most favorably.
            While logic-shrunk implementations with initial $K=5$ exhibit the greatest area savings over LUTNet, those with $K=4$ have the best area-accuracy tradeoff.
            The superiority of designs with initial $K=4$ can be attributed to the presence of 5-LUTs within those logic-shrunk from a netlist with $K=5$.
            The LUTs physically present in the target device are 6-LUTs, each capable of implementing either a single six-input function or two $k$-input functions with at least five (for $k=5$), three ($k=4$) or one ($k=3$) shared inputs.
            There is less opportunity for packing of pairs of 5-LUTs than with LUTs taking four inputs or fewer, hence the lower area efficiency of designs logic-shrunk from the starting point with $K=5$.
            We thus recommend $K=4$ as the starting point for exploration with new benchmarks.
            
        \subsubsection{Comparison to Random Pruning}
        
            \begin{figure}
                \centering
                \begin{tikzpicture}
    
    \begin{axis}[
		width=\columnwidth,
		height=0.8\columnwidth,
		ylabel near ticks,
		ylabel={Top-1 test error rate (\%)},
	    xtick scale label code/.code={\pgfmathparse{int(#1)}$\text{Area occupancy (LUTs)} \cdot 10^{\pgfmathresult}$},
        every x tick scale label/.style={at=(xticklabel cs:0.5), anchor=north},
        error bars/y dir=both,
        error bars/y explicit=true,
        error bars/error bar style={color=gray!50!white},
        xmin=100000,
        xmax=300000,
        ymax=20
    ]
        \addplot [only marks, mark=triangle, mark options={scale=1.5, color=\mycolourone!33!white}] table [y=4Lerr, x=4LLUT, y error plus=4errbarU, y error minus=4errbarL] {data/err_area_vs_rand_pruning.txt}; \label{plt:tradeoff_4lutnet_ls}
        \addplot [only marks, mark=pentagon, mark options={scale=1.5, color=\mycolourone!33!white}] table [y=4RALerr, x=4RALLUT, y error plus=4RAerrbarU, y error minus=4RAerrbarL] {data/err_area_vs_rand_pruning.txt}; \label{plt:tradeoff_4lutnet_ra}
        \addplot [very thick, \mycolourfour] table [y=4LLSerr, x=4LLSLUT] {data/pareto_4lutnet.txt};
        \addplot [very thick, \mycoloursix] table [y=4RALerr, x=4RALLUT] {data/pareto_ra_4lutnet.txt}; \label{plt:ls_4lutnet_pareto_rand_ra}
        \addplot [mark=triangle,
            mark size=4pt,
            mark options={
                draw=black,
                fill=white,
            },
            only marks,
            every mark/.append style={rotate=90},
        ]
        table {%
        105000 15.55
        };
        \node (lutnet_source) at (axis cs:282560,15.55){};
        \node (lslutnet_destination) at (axis cs:188765,15.55){};
        \draw[very thick, ->](lutnet_source)--(lslutnet_destination) node [midway, anchor=north, xshift=5mm] {1.50$\times$};
        
        \node[draw, fill=white, inner sep=0.2pt, anchor=north east] at (axis description cs:0.97, 0.96) {\small
        \begin{tabular}{cl}
            \ref{plt:tradeoff_4lutnet_ra}~\ref{plt:ls_4lutnet_pareto_rand_ra} & Randomly pruned\\
            \ref{plt:tradeoff_4lutnet_ls}~\ref{plt:ls_4lutnet_pareto} & Logic-shrunk
        \end{tabular}};

    \end{axis}

\end{tikzpicture}
            	\caption{
                	Area-accuracy tradeoff for randomly pruned~(\ref{plt:tradeoff_4lutnet_ra}) and logic-shrunk~(\ref{plt:tradeoff_4lutnet_ls}) CNV implementations trained to classify CIFAR-10 with initial LUT size $K=4$.
                	Each point reflects a distinct LUT input sparsity $\delta$.
                	Pareto frontiers for logic-shrunk~(\ref{plt:ls_4lutnet_pareto}) and randomly pruned~(\ref{plt:ls_4lutnet_pareto_rand_ra}) designs are overlaid for comparison.
                	The annotated arrow indicates the area saving between the best-performing implementations with accuracy $\pm$0.3~pp from unpruned ReBNet's~(\ref{plt:ref_acc}).
                }
            	\label{plot:AREA_LUT_TRADEOFF_VS_RAND_PRUNING}
            \end{figure}
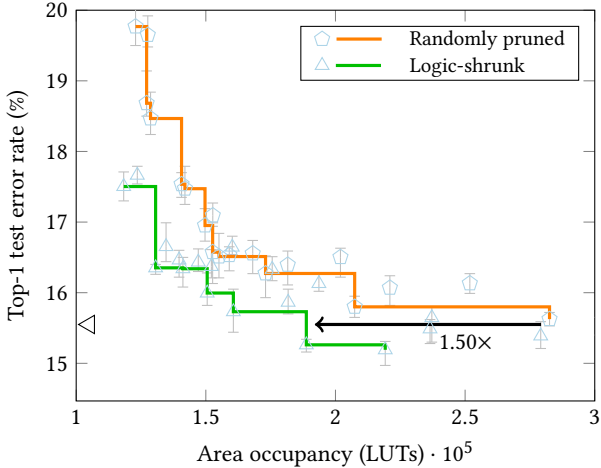
            
            To verify that logic shrinkage is an efficient sparsification method, we compared it against random LUT input pruning as a sanity check.
            The process for this was identical to that for logic shrinkage, but LUT inputs were removed at random.
            Our results for this set of experiments are shown in Figure~\ref{plot:AREA_LUT_TRADEOFF_VS_RAND_PRUNING}.
            As evidenced by their Pareto fronts, logic-shrunk~(\ref{plt:tradeoff_4lutnet_ls}~\ref{plt:ls_4lutnet_pareto}) implementations consistently outperformed those with random pruning~(\ref{plt:tradeoff_4lutnet_ra}~\ref{plt:ls_4lutnet_pareto_rand_ra}), the former achieving a 1.50$\times$ area saving {\em vs} the latter at the unpruned ReBNet accuracy~(\ref{plt:ref_acc}).
        
        \subsubsection{LUT Distribution}
        
            \begin{table*}
        		\centering
        		\caption{
        		    Pre\-/ and post-synthesis LUT size distributions for the LUTNet~(\ref{plt:tradeoff_4lutnet_pointsonly_baseline_4}) and logic-shrunk~(\ref{plt:tradeoff_4lutnet_pointsonly_curve_4}) implementations with (initial) LUT size $K=4$ and node sparsity $\theta=94\%$ reported in Figure~\ref{plot:tradeoff_4lutnet_pointsonly}.
        		    Shaded cells mark the post-synthesis LUT size in the majority.
        		}
                \begin{threeparttable}
        		    \begin{tabular}{S[table-format=2.1]S[table-format=2.2]SSSSSSSSSS}
        				\toprule
        				{\multirow{2}[1]{*}{\makecell{LUT input\\sparsity $\delta$ (\%)}}} & {\multirow{2}[1]{*}{\makecell{Top-1 test\\error rate (\%)}}} & \multicolumn{2}{c}{$\Sigma$ LUTs} & \multicolumn{2}{c}{4-LUTs} & \multicolumn{2}{c}{3-LUTs} & \multicolumn{2}{c}{2-LUTs} & \multicolumn{2}{c}{1-LUTs}\\
        				\cmidrule(lr){3-4} \cmidrule(lr){5-6} \cmidrule(lr){7-8} \cmidrule(lr){9-10} \cmidrule(lr){11-12}
        				 & & {Pre} & {Post} & {Pre} & {Post} & {Pre} & {Post} & {Pre} & {Post} & {Pre} & {Post}\\
        				\midrule
        			    0.0 & 15.89 & 70778 & 53430 & 70778 & {\cellcolor{gray!25!white}} 49541 & 0 & 3654 & 0 & 233 & 0 & 2\\
        			    25.0 & 15.49 & 70778 & 40209 & 29758 & {\cellcolor{gray!25!white}} 24405 & 20908 & 13629 & 10466 & 1852 & 9646 & 323\\
        			    50.0 & 15.18 & 70778 & 21451 & 2642 & 2293 & 18130 & {\cellcolor{gray!25!white}} 12289 & 26592 & 6375 & 23414 & 494\\
        			    75.0 & 15.26 & 62518 & 3212 & 0 & 0 & 998 & 816 & 6264 & {\cellcolor{gray!25!white}} 1708 & 55256 & 688\\
        			    87.5 & 16.00 & 35262 & 945 & 0 & 0 & 6 & 2 & 116 & 32 & 35140 & {\cellcolor{gray!25!white}} 911\\
        				\bottomrule
        			\end{tabular}
        		\end{threeparttable}
        		\label{tab:k_breakdown}
        	\end{table*}
            
            In order to better understand the source of our area savings, we inspected the post-shrinkage distribution of LUT sizes $K^\prime_n$ for each LUT $n$ in both pre\-/ and post-synthesis netlists.
            To facilitate our investigation, we disabled design hierarchy optimization in Vivado, preventing the synthesis engine from flattening across modules.
            Table~\ref{tab:k_breakdown} shows the breakdown in LUT sizes across the implementations shown in Figure~\ref{plot:AREA_LUT_TRADEOFF_GROUPPLOT_POINTS_ONLY} with (initial) LUT size $K=4$ and node sparsity $\theta=94.0\%$~(\ref{plt:tradeoff_2lutnet_pointsonly_baseline_4}~\ref{plt:tradeoff_2lutnet_pointsonly_curve_4}) as an example.
            The implementation with LUT input sparsity $\delta=0$ is the LUTNet design; all of those with $\delta>0$ were logic-shrunk from that starting point.
            Pre-synthesis netlists were those generated as output from the logic shrinkage (or vanilla LUTNet) toolflow, while post-synthesis netlists were extracted from Vivado before implementation.
            
            Two key features are apparent from the data in Table~\ref{tab:k_breakdown}.
            Firstly, there is a downward (towards high sparsity) and rightward (small LUTs) shift in LUT counts.
            Diminishing returns are seen when increasing $K$ in LUTNet architectures~\cite{LUTNET}, indicating that the inputs added with higher $K$ tend to be of decreasing value.
            These are generally severed first, making it increasingly unlikely that all inputs of large LUTs will remain unpruned as $\delta$ rises.
            As a result, we see that larger LUTs are usually reduced in size before smaller ones, giving rise to the reduction in majority LUT size with increasing $\delta$ highlighted with shading.
            We can also infer from these data, along with reference back to Figure~\ref{plot:AREA_LUT_TRADEOFF_GROUPPLOT_POINTS_ONLY}, that equally sparse designs perform better under logic shrinkage than when constructed using the vanilla LUTNet flow.
            For the same $\theta$, logic shrinkage with initial $K=4$ and $\delta=0.5$ generates a netlist with the same number of total LUT inputs as a LUTNet design with $K=2$.
            However, the logic-shrunk implementation has an error rate of 15.18\% (Table~\ref{tab:k_breakdown}): lower than all LUTNet designs with $K=2$ (Figure~\ref{plot:AREA_LUT_TRADEOFF_GROUPPLOT_POINTS_ONLY}).
            It is thus evident that selectively shrinking to a smaller implementation from a larger one through consideration of LUT input salience is preferable to the creation of an equally sized architecture from scratch.
            
            Secondly, there are large gaps between pre\-/ and post-synthesis LUT counts, with this phenomenon becoming more pronounced as $\delta$ increases.
            This is attributable to the logic optimization central to synthesis, opportunities for which increase as LUT size falls.
            The effects of optimization are particularly marked for 1-LUTs, the majority of which were optimized away.
            Three of the four possible functions performable by a 1-LUT ($y=0$, $y=1$, $y=x$) are free to implement.
            Only $y=\overline{x}$ requires device resources, but in most cases can be absorbed by the downstream logic.
            Consequently, we see increasing LUT removal as the average LUT size decreases.
            Overall, we can conclude that logic shrinkage successfully promotes sparsity in such a way as to suit the optimizations performed during synthesis, resulting in highly area-efficient implementations.
    
        \subsubsection{Other Benchmarks}

        	\begin{table*}
        		\centering
        		\caption{
        		    Top-1 test error rate and area---post-synthesis and post-implementation---for LUTNet and logic-shrunk designs with various models classifying various datasets. 
        		    (Initial) LUT size $K$ was 4 in all cases.
        		}
                \begin{threeparttable}
        		    \begin{tabular}{ccS[table-format=2.1]S[table-format=2.1]S[table-format=2.2]S[table-format=1.2]S[table-format=7]S[table-format=1.2]S[table-format=6]S[table-format=1.2]}
        				\toprule
        				\multirow{2}[1]{*}{\makecell{Dataset\\(network)}} & \multirow{2}[1]{*}{Architecture} & {\multirow{2}[1]{*}{\makecell{Node sparsity\\$\theta$ (\%)}}} & {\multirow{2}[1]{*}{\makecell{LUT input\\sparsity $\delta$ (\%)}}} & \multicolumn{2}{c}{Error rate} & \multicolumn{2}{c}{Area (post-synth.)} & \multicolumn{2}{c}{Area (post-impl.)}\\
        				\cmidrule(lr){5-6}\cmidrule(lr){7-8}\cmidrule(lr){9-10}
        				 &  &  &  & {\%} & {$\Delta$ (pp)} & {LUTs} & {$\Delta$ ($\times\downarrow$)} & {LUTs} & {$\Delta$ ($\times\downarrow$)}\\
        				\midrule
        				\multirow{2}{*}{\makecell{MNIST\\(LFC)}} & LUTNet & 99.9 & {--} & 2.13 & {--} & 62919 & {--} & 58192 & {--}\\
        				 & Logic-shrunk & 90.0 & 75.0 & 2.53 & 0.40 & 63928 & 0.98
        				 & 54647 & 1.06\\
        				\midrule
        				\multirow{2}{*}{\makecell{SVHN\\(CNV)}} & LUTNet & 95.0 & {--} & 3.80 & {--} & 201644 & {--} & 154814 & {--}\\
        				 & Logic-shrunk & 98.0 & 75.0 & 3.75 & -0.05 & 179236 & 1.13 & 137610 & 1.13\\
        				\midrule
        				\multirow{2}{*}{\makecell{CIFAR-10\\(CNV)}} & LUTNet & 91.0 & {--} & 15.42 & {--} & 339479 & {--} & 291349 & {--}\\
        				 & Logic-shrunk & 94.0 & 75.0 & 15.26 & -0.16 & 220060 & 1.54 & 188765 & 1.54\\
        				\midrule
        				\multirow{2}{*}{\makecell{ImageNet\\(Bi-Real-18)\tnote{1}}} & LUTNet & 30.0 & {--} & 45.13 & {--} & 1840666 & {--} & {--}\tnote{2} & {--}\\
        				 & Logic-shrunk & 30.0 & 75.0 & 46.60 & 1.47 & 690357 & 2.67 & 665720 & {--}\\
        				\bottomrule
        			\end{tabular}
        			\begin{tablenotes}
        			    \item[1] Target layer only. Designs for other datasets included all network layers.
        			    \item[2] Design could not fit onto target device.
        			\end{tablenotes}
        		\end{threeparttable}
        		\label{tab:all_datasets}
        	\end{table*}
            
            We also benchmarked logic shrinkage using other popular datasets and models: MNIST (with LFC), SVHN (with CNV) and ImageNet (with Bi-Real-18).
            Table~\ref{tab:all_datasets} shows the post-synthesis and post-implementation LUT requirements of each of these model-dataset combinations when implemented with LUTNet and logic-shrunk architectures with (initial) LUT size $K=4$.
            The same layers for all pairs of designs were unrolled and pruned, with the node sparsity (and LUT input sparsity) tuned in an effort to keep their accuracy as close as possible.
                
            For CNV classifying CIFAR-10, our use of logic shrinkage saw an area reduction of 1.54$\times$.
            With the smaller datasets, the gains realized via logic shrinkage were less pronounced.
            The SVHN-CNV and MNIST-LFC combinations are more tolerant of sparsity, thus the majority of nodes in these networks were able to be removed prior to logic expansion.
            This left relatively little room for further improvement by logic shrinkage.
            Despite this, we still achieved area reductions of around 10\% for these simpler tasks.
            For ImageNet on Bi-Real-18, the LUTNet layer was too large to fit our target FPGA, the XCVU9P (1182240 LUTs).
            Logic shrinkage with node and LUT input sparsity of 30\% and 75\%, respectively, saw its post-synthesis area reduced by 2.67$\times$, thus leading to success in implementation.
            

    \subsection{Energy Efficiency}

        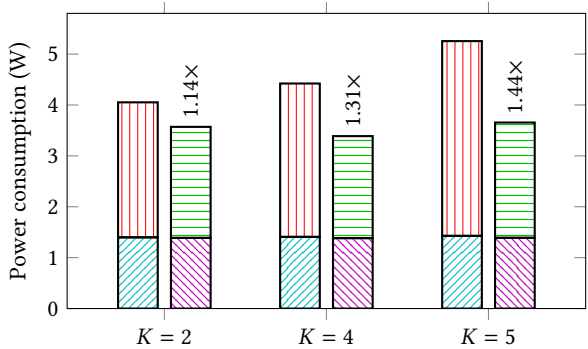
\begin{figure}
        	\begin{tikzpicture}

    \begin{axis}[
        ybar stacked,
        height=0.65\columnwidth,
        width=\columnwidth,
		enlarge x limits=0.3,
		max space between ticks=20,
        ymin=0,
        ymax=5.8,
        ylabel near ticks,
		ylabel={Power consumption (W)},
        xtick=data,
        xticklabels from table={data/energy_efficiency.txt}{Name},
        xtick align=outside,
        /pgf/bar width=15pt,
        legend image code/.code={
            \draw[#1, bar width=6pt, yshift=-0.3em] plot coordinates {(0cm,0.8em)};
        }
    ]
        
        \addplot [bar shift=-10pt, thick, pattern=\mypatternone, pattern color=\mycolourone] table [x=id, y=static] {data/energy_efficiency.txt};
        \label{plt:static_power}
        \addplot [bar shift=-10pt, thick, pattern=\mypatternthree, pattern color=\mycolourthree] table [x=id, y=dynamic] {data/energy_efficiency.txt};
        \label{plt:dynamic_power}

    \end{axis}
    
    \begin{axis}[
        ybar stacked,
        height=0.65\columnwidth,
        width=\columnwidth,
		enlarge x limits=0.3,
        ymin=0,
        ymax=5.8,
        axis x line=none,
        axis y line=none,
		xtick=\empty,
		ytick=\empty,
        /pgf/bar width=15pt,
        legend image code/.code={
            \draw[#1, bar width=6pt, yshift=-0.3em] plot coordinates {(0cm,0.8em)};
        }
    ]
        
        \addplot [bar shift=10pt, thick, pattern=\mypatterntwo, pattern color=\mycolourtwo] table [x=id, y=static] {data/energy_efficiency_ls.txt};
        \label{plt:static_power_ls}
        \addplot [bar shift=10pt, thick, pattern=\mypatternfour, pattern color=\mycolourfour] table [x=id, y=dynamic] {data/energy_efficiency_ls.txt}
        node [pos=0, rotate=90, anchor=west, yshift=-10pt] {1.14$\times$}
        node [pos=0.5, rotate=90, anchor=west, yshift=-10pt] {1.31$\times$}
        node [pos=1, rotate=90, anchor=west, yshift=-10pt] {1.44$\times$};
        \label{plt:dynamic_power_ls}
    
    \end{axis}

\end{tikzpicture}
        	\caption{
                Post-implementation power consumption estimates for the LUTNet~(\ref{plt:static_power}~\ref{plt:dynamic_power}) and logic-shrunk~(\ref{plt:static_power_ls}~\ref{plt:dynamic_power_ls}) designs in Figure~\ref{plot:area_savings}.
                Power is broken into static~(\ref{plt:static_power}~\ref{plt:static_power_ls}) and dynamic~(\ref{plt:dynamic_power}~\ref{plt:dynamic_power_ls}) components.
                Annotations reflect the decrease in total power between each pair of implementations.
            }
        	\label{plot:energy_efficiency}
        \end{figure}
        
        We also sought to quantify the energy efficiency impact attributable to logic shrinkage.
        To do so, we obtained power consumption estimates of both LUTNet and logic-shrunk implementations using the Xilinx Power Analyzer (XPA) tool with default settings: vectorless mode and 12.5\% primary input switching probability.
        The resultant power estimates, for the same designs as captured in Figure~\ref{plot:area_savings}, are shown in Figure~\ref{plot:energy_efficiency}.
        All were obtained post-placement and \-/routing.
        Power consumption is equivalent to energy efficiency here since all implementations have identical throughput.
        
        Since dynamic power consumption is directly related to area occupancy, Figures~\ref{plot:area_savings} and \ref{plot:energy_efficiency} show similar trends.
        The static power remains consistent across all implementations.
        Overall, it can be concluded that the significant area reductions of logic shrinkage also result in energy efficiency improvements.
            
    \subsection{Training Efficiency}
        
        \begin{figure}
            \centering
            \begin{tikzpicture}
    
    \begin{axis}[
        ybar,
        height=0.65\columnwidth,
        width=\columnwidth,
		enlarge x limits=0.3,
        ymin=1,
        ymax=1.2,
        ylabel near ticks,
		ylabel={$\Delta$ training time/epoch ($\times\uparrow$)},
        xtick=data,
        xticklabels from table={data/training_speed.txt}{Name},
        xtick align=outside,
        /pgf/bar width=15pt,
        legend image code/.code={
            \draw[#1, bar width=6pt, yshift=-0.3em] plot coordinates {(0cm,0.8em)};
        }
    ]
        
        \addplot [bar shift=0pt, thick, pattern=\mypatternone, pattern color=\mycolourone] table [x=id, y=time_change] {data/training_speed.txt};
        \label{plt:training_speed_baseline}

    \end{axis}
    
        
    

\end{tikzpicture}
        	\caption{Training time increase for the logic-shrunk designs over LUTNet taken from Table~\ref{tab:all_datasets}.}
        	\label{plot:training_speed}
        \end{figure}
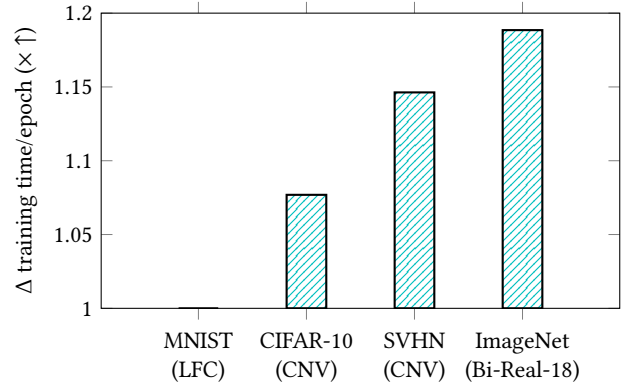
        
        Logic shrinkage introduces additional matrix-vector multiplications for every forward propagation during retraining in order to ensure that pruned inputs remain severed.
        Thanks to the highly optimized linear algebra routines provided by GPUs, the slowdown in training speed with logic shrinkage is minor.
        This is evident in Figure~\ref{plot:training_speed}, in which we capture per-epoch logic shrinkage overheads.
    
\section{Conclusion}

    In this paper, we introduced logic shrinkage: the automated search for, and implementation of, LUT-based neural network inference accelerators in which LUT sizes and inputs are learned during training.
    We showed our realization of logic shrinkage to be lightweight and to result in the production of netlists that well suit the logic optimizations performed by FPGA synthesis tools.
    We analyzed hundreds of experimental results, finding significant area-accuracy tradeoff improvement over homogeneous LUT-based networks.
    
    The authors of prior NAS works pursued a top-down approach, learning an intermediate representation---a network topology---while leaving its hardware mapping as a separate task.
    In contrast, we propose a bottom-up, hardware-aware alternative: directly learning a netlist as the topology, with the flexibility of the target platform exposed to the training process.
    We chose to focus on the search for efficient node functions in this work, but in the future we will extend our scope to other components---accumulators, activation functions, {\em etc.}---in order to allow for greater structural learning by the training software and to further drive up the area and energy efficiency of the resulting inference engines.

\section*{Acknowledgments}

    The authors are grateful for the support of the United Kingdom EPSRC (grant numbers EP/S030069/1 and EP/P010040/1).

\newpage
\bibliographystyle{ACM-Reference-Format}
\bibliography{bibliography}

\end{document}